\documentclass[11pt]{article}

\usepackage[preprint]{acl}

\usepackage{times}
\usepackage{latexsym}

\usepackage[T1]{fontenc}

\usepackage[utf8]{inputenc}

\usepackage{microtype}

\usepackage{inconsolata}

\usepackage{graphicx}

\usepackage{amsmath}
\usepackage{multirow}

\usepackage{float}
\usepackage{tikz}
\usepackage{lengthconvert} 
\usetikzlibrary{patterns}   

\def\thickhline{\noalign{\hrule height.8pt}}

\title{RBCorr: Response Bias Correction in Language Models}

\author{Om Bhatt \\
  Cognitive Sciences \\ University of California, Irvine \\
  \texttt{om.bhatt@uci.edu} \\\And
  Anna A. Ivanova \\
  School of Psychology \\ Georgia Institute of Technology \\
  \texttt{a.ivanova@gatech.edu} \\}

\usepackage[indent=1em]{parskip}

\begin{document}
\maketitle
\begin{abstract}

Language models (LMs) are known to be prone to response biases, which present as option preference biases in fixed-response questions. It is therefore imperative to develop low-cost and effective response bias correction methods to improve LM performance and enable more accurate evaluations of model abilities. Here, we propose a simple response bias correction strategy (\texttt{RBCorr}) and test it on 12 open-weight language models using yes-no, entailment, and multiple choice questions. We show that response bias is prevalent in LMs pre-correction and that \texttt{RBCorr} effectively eliminates bias and boosts model performance. We also explore the generalizability of bias behavior across models, datasets, and prompt formats, showing that LogProbs-based correction is highly dependent on all three of these aspects. Overall, \texttt{RBCorr} is an easy-to-use method that can boost the performance of smaller LMs and ensure that LM performance on closed-response benchmarks aligns more closely with their true capabilities. Code: \url{https://github.com/ombbhatt/rbcorr_bias_correction}
\end{abstract}

\section{Introduction}

In fixed-response evaluation settings, language models (LMs) have been reported to suffer from response bias \citep[e.g., ][]{pezesh2023-optionorderingbias, tjuatja2024-llmhumanlikebias, contextcalibpaper}, which reduces their performance and viability as real-world reasoners. Recent mitigation efforts have looked towards LogProbs manipulations as a general strategy for its ease of accessibility and low cost, but we believe that further advancement also requires a systematic exploration of the experimental setup (i.e. the model, dataset, and prompt scheme being used), as well as tracking bias separately from accuracy to gain more nuanced insights on the effects of applying correction, and finally, new manipulation strategies themselves to try and improve upon existing methods.

This paper investigates and corrects response biases in open-weight LMs for fixed-response questions. We test LMs across a variety of reasoning tasks and prompt formats, track both bias and accuracy, and propose a simple LogProbs-based correction method that can be applied to multiple question types, effectively mitigating bias and increasing or maintaining accuracy. Additionally, we compare our method to existing correction methods, characterize models' baseline label preferences, and perform further analysis to reveal that LogProbs values are highly context-specific to the model, dataset, and prompt formats being used. Our results show that a straightforward calibration-based bias correction strategy, akin to approaches in traditional machine learning \citep[e.g., ][]{saerens2002adjusting, zadrozny2002transforming}, can be effectively applied to today's LMs to improve their performance and enable more faithful evaluations of their capabilities.
 
\section{Approach}

Our approach is based on extracting token log probabilities (LogProbs) from the last layer of an LM, prior to softmax transformation and final token sampling. In a bias-free model, the average LogProbs for different response options should be equal (as long as the dataset is class-balanced). Our goal is to measure the deviation between this uniform distribution and empirically observed model LogProbs scores (i.e., response bias) and to correct it by applying a correction term to an LM's LogProbs values prior to response generation. 

\subsection{Extracting Model Response}

For each type of question, we define single-token response formats in the prompts. We then extract the log-probability values for the appropriate set of response tokens for each dataset item depending on the question type (we test on 2-choice \texttt{`Yes'/`No'} multi-domain questions, 3-choice \texttt{`0'/`1'/`2'} (neutrality/entailment/contradiction) NLI questions, and 4-choice \texttt{`A'/`B'/`C'/`D'} multi-domain questions). The token with the highest log-probability value is recorded as the model's response. To account for response variation, we consider the whitespace-prepended version of each token as well (we log-sum-exp the LogProbs values for \texttt{`\textvisiblespace Yes'} and `\texttt{Yes}' when evaluating 2-choice questions, for example.) 

\subsection{Measuring Response Bias}

We use the \textbf{Total Variation Distance (TVD)} value between the ground-truth label distribution in the test dataset and the model response distribution to measure model bias. TVD yields a value between 0 and 1, where 0 indicates identical distributions.

Our test datasets are class-balanced, which means that the ground-truth label distribution is a uniform distribution, and we can anchor it as the point of comparison of what an ideal unbiased model output distribution should be. We can then calculate the TVD between the uniform distribution and the actual model response distribution, both before and after applying a corrective LogProbs manipulation, and finally compare the resulting TVD values. We can infer whether biased behavior was reduced by checking for a lower TVD value yielded using the corrected response distribution.

For a ground-truth dataset label distribution $G$ (uniform) and model response label distribution $M$, we measure model bias as the TVD between them:
\begin{equation}
\text{TVD}(G, M) = \frac{1}{2} \sum_{x \in X} |G(x) - M(x)|
\end{equation}
Where $X$ denotes the relevant option label space. This way of measuring bias is a simplified version of the method described by \citet{tvdmetricpaper}, where they use an unbalanced input dataset, but sample a uniform distribution of model output probabilities before calculating the TVD.
\subsection{Applying RBCorr Correction}

Our correction method, \texttt{RBCorr}, applies a mean-normalization to individual item LogProbs values by using a small held-out class-balanced calibration set for mean LogProbs estimation for each response option. To perform this correction, we first extract model responses and the options' LogProbs values for each item in a dataset. We then sample a small calibration set of questions from the entire dataset, and calculate the mean of the LogProbs value for each of the option tokens in that set. Importantly, we enforce class-balance when sampling the calibration set, to ensure that the means estimation is not skewed by over- or under-representation of any class of questions. We finally subtract these means from the corresponding tokens' LogProbs for every item in the evaluation set, i.e., all questions that were not part of the calibration set. We extract model responses from this corrected set of LogProbs to record our `correction-applied' results. This correction requires no overhead computation to perform since it only requires the baseline LogProbs values for adjustment.

\begin{figure*}[t]
    \includegraphics[width=\linewidth]
    {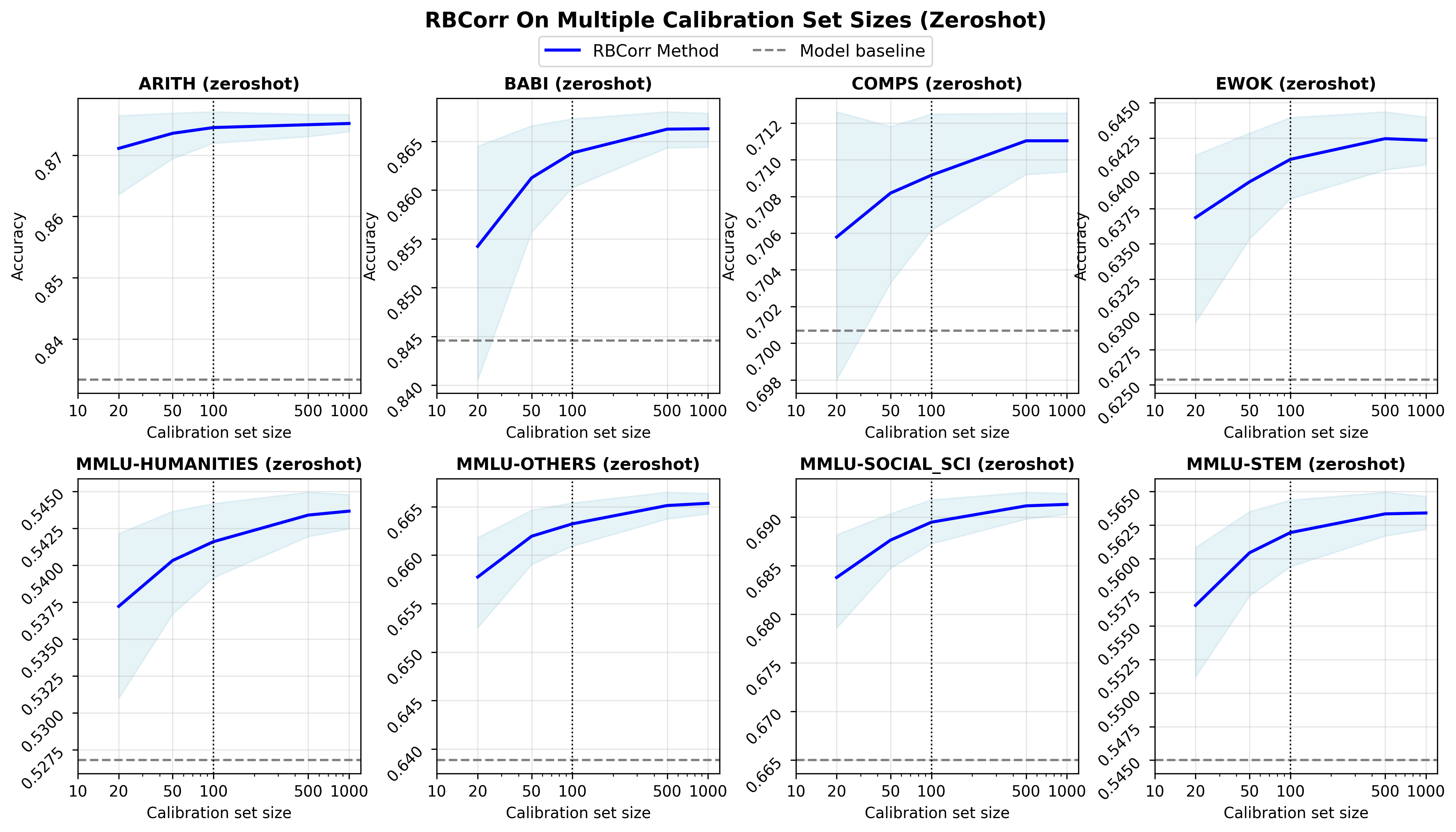}
    \caption{Accuracy achieved by applying RBCorr at multiple calibration set sizes, averaged across all models. Shading shows the interquartile range of accuracies achieved after performing 100 separate iterations of the correction process on each dataset. Horizontal dashed line shows average baseline accuracy across all models. We discuss our method's results achieved using a set size of 100 (marked with vertical line) as a realistic setup.}
    \label{fig:rbcorrsizes}
\end{figure*}

\textbf{How big does the calibration set need to be?} This question is especially relevant in real-world deployment, where datasets may be only partially labeled or accessible during online inference. To explore this, we perform our correction method using a range of fixed calibration set sizes: 20, 50, 100, 500, and 1000 questions. For each size, we run 100 iterations of this correction procedure using different randomly-sampled calibration sets every time to get a robust idea of how reliably the values are adjusted with random calibration set selection, and plot the mean accuracy gain across all runs along with the interquartile range (see Figure \ref{fig:rbcorrsizes}). We find that the accuracy gains only increase marginally with with bigger set sizes --- using 100 calibration questions generally provides stable correction effects, which also highlights the low-data requirement to use our method. Later experiments consequently discuss our method's efficacy using a calibration set size of 100.

\begin{figure*}[t]
    \centering
    \includegraphics[width=\linewidth]{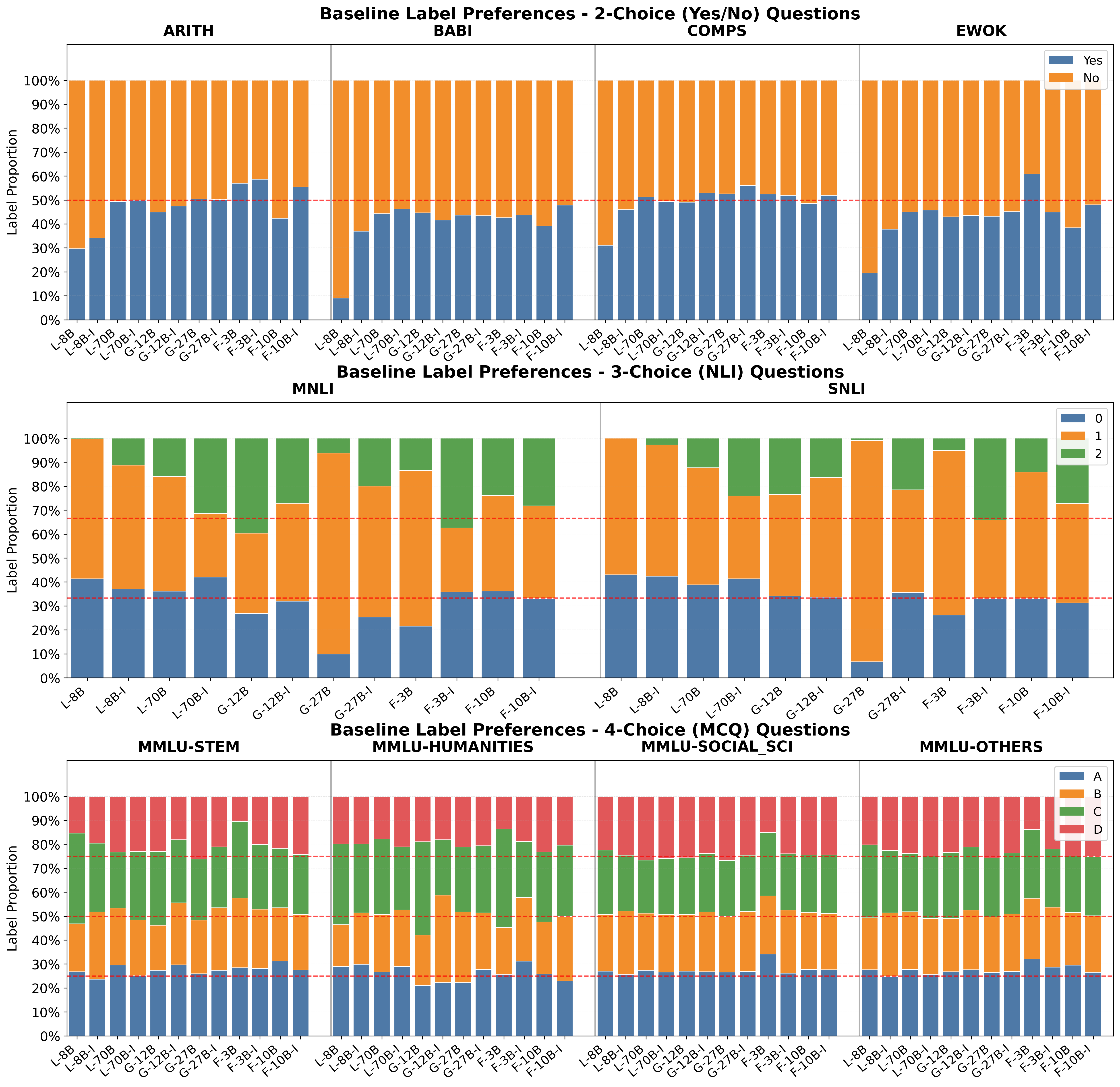}
    \caption{Baseline model response label distribution for all models on all datasets, using the fewshot prompt format. Red dotted lines indicate uniform distribution intervals (i.e. the dataset's ground-truth label distribution.)}
    \label{fig:baselinemodelbiasbarchart}
\end{figure*}

\begin{figure*}[t]
    \centering
    \includegraphics[width=\linewidth]{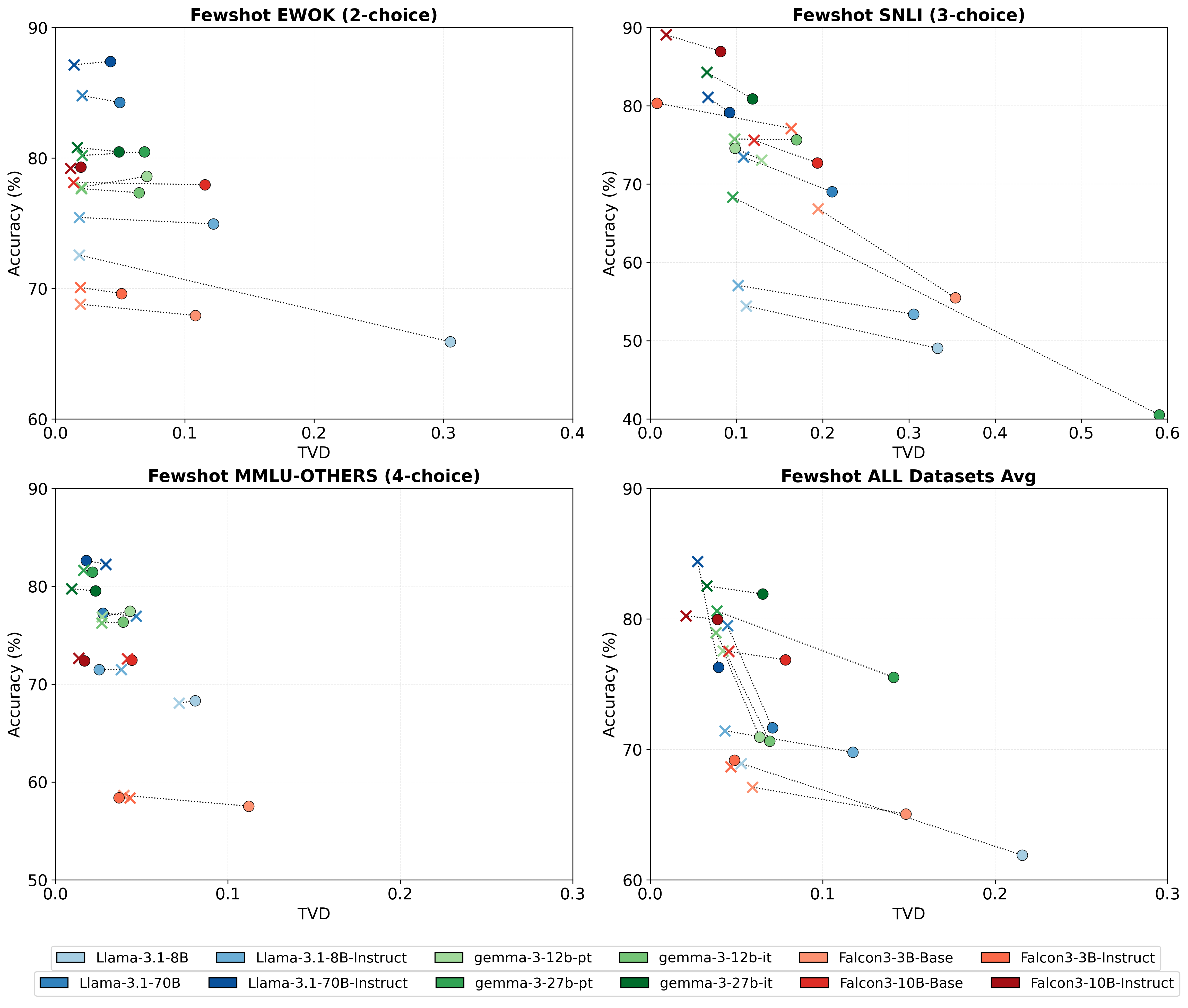}
    \caption{Scatterplots showing per-model bias (TVD; $\downarrow$ is better) and accuracy (\%; $\uparrow$ is better) before [$\bullet$] vs. after [$\times$] applying \texttt{RBCorr} correction. We show results on one dataset per each question-type; the bottom-right plot shows results averaged across all ten datasets.}
    \label{fig:samecondscatterplots}
\end{figure*}

\begin{table*}[t]
\small
\setlength{\tabcolsep}{9.14pt}
\renewcommand{\arraystretch}{1.5}
\begin{tabular}{|c|c|cc|cc|cc|cc|}
\hline
Dataset & \begin{tabular}[c]{@{}c@{}}Model\\ (Llama3.1)\end{tabular} & \multicolumn{2}{c|}{\begin{tabular}[c]{@{}c@{}}Baseline\\ (Acc | Bias)\end{tabular}} & \multicolumn{2}{c|}{\begin{tabular}[c]{@{}c@{}}CC\\ $\Delta$(Acc | Bias)\end{tabular}} & \multicolumn{2}{c|}{\begin{tabular}[c]{@{}c@{}}BC\\ $\Delta$(Acc | Bias)\end{tabular}} & \multicolumn{2}{c|}{\begin{tabular}[c]{@{}c@{}}RBCorr (ours)\\ $\Delta$(Acc | Bias)\end{tabular}} \\ \hline
\multirow{2}{*}{\begin{tabular}[c]{@{}c@{}}ARITH\\ (2-Choice)\end{tabular}} & 8B & \multicolumn{1}{c|}{74.2} & 0.203 & \multicolumn{1}{c|}{+11.8} & -0.182 & \multicolumn{1}{c|}{+12.1} & -0.086 & \multicolumn{1}{c|}{\textbf{+12.4}} & \textit{-0.194} \\ \cline{2-10} 
 & 8B-Instruct & \multicolumn{1}{c|}{78.8} & 0.159 & \multicolumn{1}{c|}{+3.2} & -0.058 & \multicolumn{1}{c|}{+5.2} & -0.090 & \multicolumn{1}{c|}{\textbf{+5.8}} & \textit{-0.151} \\ \hline
\multirow{2}{*}{\begin{tabular}[c]{@{}c@{}}BABI\\ (2-Choice)\end{tabular}} & 8B & \multicolumn{1}{c|}{59.0} & 0.410 & \multicolumn{1}{c|}{+11.8} & -0.123 & \multicolumn{1}{c|}{+26.5} & -0.307 & \multicolumn{1}{c|}{\textbf{+28.9}} & \textit{-0.401} \\ \cline{2-10} 
 & 8B-Instruct & \multicolumn{1}{c|}{86.6} & 0.131 & \multicolumn{1}{c|}{\textbf{+2.8}} & -0.040 & \multicolumn{1}{c|}{+2.1} & -0.114 & \multicolumn{1}{c|}{+2.1} & \textit{-0.120} \\ \hline
\multirow{2}{*}{\begin{tabular}[c]{@{}c@{}}COMPS\\ (2-Choice)\end{tabular}} & 8B & \multicolumn{1}{c|}{79.0} & 0.185 & \multicolumn{1}{c|}{+6.9} & -0.098 & \multicolumn{1}{c|}{\textbf{+8.5}} & -0.173 & \multicolumn{1}{c|}{+8.0} & \textit{-0.174} \\ \cline{2-10} 
 & 8B-Instruct & \multicolumn{1}{c|}{86.3} & 0.036 & \multicolumn{1}{c|}{-0.1} & -0.016 & \multicolumn{1}{c|}{\textbf{0.0}} & -0.003 & \multicolumn{1}{c|}{\textbf{0.0}} & \textit{-0.025} \\ \hline
\multirow{2}{*}{\begin{tabular}[c]{@{}c@{}}EWOK\\ (2-Choice)\end{tabular}} & 8B & \multicolumn{1}{c|}{65.9} & 0.305 & \multicolumn{1}{c|}{\textbf{+6.6}} & -0.280 & \multicolumn{1}{c|}{+6.0} & -0.221 & \multicolumn{1}{c|}{\textbf{+6.6}} & \textit{-0.287} \\ \cline{2-10} 
 & 8B-Instruct & \multicolumn{1}{c|}{75.0} & 0.122 & \multicolumn{1}{c|}{\textbf{+0.7}} & -0.087 & \multicolumn{1}{c|}{-0.7} & -0.037 & \multicolumn{1}{c|}{+0.5} & \textit{-0.104} \\ \thickhline
\multirow{2}{*}{\begin{tabular}[c]{@{}c@{}}SNLI\\ (3-Choice)\end{tabular}} & 8B & \multicolumn{1}{c|}{49.0} & 0.333 & \multicolumn{1}{c|}{\textbf{+11.7}} & -0.029 & \multicolumn{1}{c|}{+4.3} & -0.143 & \multicolumn{1}{c|}{+5.4} & \textit{-0.222} \\ \cline{2-10} 
 & 8B-Instruct & \multicolumn{1}{c|}{53.4} & 0.305 & \multicolumn{1}{c|}{+1.2} & -0.095 & \multicolumn{1}{c|}{+2.2} & -0.082 & \multicolumn{1}{c|}{\textbf{+3.7}} & \textit{-0.204} \\ \hline
\multirow{2}{*}{\begin{tabular}[c]{@{}c@{}}MNLI\\ (3-Choice)\end{tabular}} & 8B & \multicolumn{1}{c|}{49.2} & 0.331 & \multicolumn{1}{c|}{+8.7} & +0.001 & \multicolumn{1}{c|}{+8.8} & -0.140 & \multicolumn{1}{c|}{\textbf{+9.9}} & \textit{-0.214} \\ \cline{2-10} 
 & 8B-Instruct & \multicolumn{1}{c|}{59.7} & 0.221 & \multicolumn{1}{c|}{+4.4} & -0.017 & \multicolumn{1}{c|}{+3.6} & -0.076 & \multicolumn{1}{c|}{\textbf{+4.7}} & \textit{-0.104} \\ \thickhline
\multirow{2}{*}{\begin{tabular}[c]{@{}c@{}}HUMANITIES\\ (4-Choice)\end{tabular}} & 8B & \multicolumn{1}{c|}{52.5} & 0.125 & \multicolumn{1}{c|}{-10.0} & +0.409 & \multicolumn{1}{c|}{\textbf{-0.4}} & +0.018 & \multicolumn{1}{c|}{-0.7} & \textit{-0.061} \\ \cline{2-10} 
 & 8B-Instruct & \multicolumn{1}{c|}{57.6} & 0.086 & \multicolumn{1}{c|}{-0.3} & +0.068 & \multicolumn{1}{c|}{\textbf{0.0}} & \textit{-0.047} & \multicolumn{1}{c|}{-0.2} & -0.031 \\ \hline
\multirow{2}{*}{\begin{tabular}[c]{@{}c@{}}OTHERS\\ (4-Choice)\end{tabular}} & 8B & \multicolumn{1}{c|}{68.3} & 0.081 & \multicolumn{1}{c|}{-9.5} & +0.259 & \multicolumn{1}{c|}{-0.7} & +0.007 & \multicolumn{1}{c|}{\textbf{-0.3}} & \textit{-0.009} \\ \cline{2-10} 
 & 8B-Instruct & \multicolumn{1}{c|}{71.5} & 0.025 & \multicolumn{1}{c|}{-0.7} & +0.073 & \multicolumn{1}{c|}{\textbf{0.0}} & \textit{+0.004} & \multicolumn{1}{c|}{\textbf{0.0}} & +0.013 \\ \hline
\multirow{2}{*}{\begin{tabular}[c]{@{}c@{}}SOCIAL SCI.\\ (4-Choice)\end{tabular}} & 8B & \multicolumn{1}{c|}{69.7} & 0.038 & \multicolumn{1}{c|}{-14.0} & +0.348 & \multicolumn{1}{c|}{\textbf{0.0}} & \textit{+0.037} & \multicolumn{1}{c|}{-0.3} & +0.040 \\ \cline{2-10} 
 & 8B-Instruct & \multicolumn{1}{c|}{72.6} & 0.021 & \multicolumn{1}{c|}{-1.0} & +0.090 & \multicolumn{1}{c|}{\textbf{+0.5}} & \textit{-0.006} & \multicolumn{1}{c|}{+0.3} & +0.004 \\ \hline
\multirow{2}{*}{\begin{tabular}[c]{@{}c@{}}STEM\\ (4-Choice)\end{tabular}} & 8B & \multicolumn{1}{c|}{52.1} & 0.146 & \multicolumn{1}{c|}{-10.5} & +0.370 & \multicolumn{1}{c|}{\textbf{+0.3}} & -0.033 & \multicolumn{1}{c|}{+0.2} & \textit{-0.108} \\ \cline{2-10} 
 & 8B-Instruct & \multicolumn{1}{c|}{56.5} & 0.068 & \multicolumn{1}{c|}{-1.7} & +0.111 & \multicolumn{1}{c|}{\textbf{-0.2}} & \textit{-0.033} & \multicolumn{1}{c|}{-0.4} & -0.020 \\ \hline
\end{tabular}
\caption{Comparison between Contextual Calibration (CC), Batch Calibration (BC) and our method (RBCorr), on all datasets using the small Llama3.1 model pair and fewshot prompt format. Best accuracy change (highest $\uparrow$ / lowest $\downarrow$) is shown in bold and best TVD change (highest $\downarrow$ / lowest $\uparrow$) is italicized in each row.}
\label{tab:methodcomparetable}
\end{table*}

\begin{table*}[t]
\small
\setlength{\tabcolsep}{10.3pt}
\renewcommand{\arraystretch}{1.5}
\begin{tabular}{|c|c|c|c|c|c|c|}
\hline
Modality & Total Pairs & Succ. Transfers & \begin{tabular}[c]{@{}c@{}}Avg $\Delta$Acc\\ (success)\end{tabular} & \begin{tabular}[c]{@{}c@{}}Avg $\Delta$TVD\\ (success)\end{tabular} &
\begin{tabular}[c]{@{}c@{}}Avg $\Delta$Acc\\ (all pairs)\end{tabular} &
\begin{tabular}[c]{@{}c@{}}Avg $\Delta$TVD\\ (all pairs)\end{tabular}\\ \hline
Cross-Dataset & 912 & 200 (\textcolor{red}{21.93\%}) & 0.0379 & 0.1175 & 0.0130 & 0.0247 \\ \hline
Cross-Model & 1008 & 105 (\textcolor{red}{10.42\%}) & 0.0619 & 0.0647 & -0.0006 & -0.0295 \\ \hline
Cross-Prompt & 624 & 65 (\textcolor{red}{10.42\%}) & 0.0174 & 0.0612 & -0.0002 & -0.0136 \\ \hline
\end{tabular}
\caption{Quantifying successful transfer of \texttt{RBCorr}'s bias correction term across all three modalities for all valid pairs of source and target in our complete experimental setup. Successful transfer correction is defined as $\geq$80\% preservation of bias reduction and accuracy increase compared to same-condition correction.}
\label{tab:transfertable}
\end{table*}

\section{Experimental Setup}

Here we describe the datasets, models, and prompt formats used to test our correction method and perform subsequent experiments for further analysis. 

\subsection{Datasets}

We test LMs on datasets with two, three and four answer options, with the aim of covering a variety of question types and knowledge domains.

\textbf{2-Choice:\quad} We use subsets of four existing datasets converted into Yes-No format, totaling to 6600 questions: \textsl{\textbf{(1) ARITH}}, a collection of 1200 addition and subtraction problems, sampled from the Arithmetic subset of the BIGBench dataset \citep{arithpaper} (e.g., \texttt{"Is 2 plus 4 equal to \{4 / 6\}?"}); \textsl{\textbf{(2) bAbI}}, a collection of 1200 questions testing 12 reading comprehension tasks, derived from the bAbI dataset \citep{babipaper} (e.g., \texttt{"Julie travelled to the park. Is Julie in the \{bedroom / park\}?"} tests basic coreference); \textsl{\textbf{(3) COMPS}}, a collection of 2100 questions testing basic property inheritance in minimal pairs (e.g., \texttt{"Does \{an iguana / a trolley\} bask in the sun?"}), derived from the COMPS dataset \citep{compspaper}; and \textsl{\textbf{(4) EWoK}}, a collection of 2136 questions testing contextual knowledge across 11 domains with a nested minimal pair construction (e.g., \texttt{"Chao is making Yan's job \{easier / harder\}. Is Chao \{helping / hindering\} Yan?"} tests social interaction), derived from the EWoK dataset \citep{ewokpaper}.

\textbf{3-Choice:\quad} We use the \textsl{\textbf{SNLI}} \citep{snlipaper} and \textsl{\textbf{MultiNLI}} \citep{mnlipaper} datasets. Both datasets consist of Recognizing Textual Entailment (RTE) questions, where a premise and a hypothesis sentence are provided, and the task is to classify whether the hypothesis entails, contradicts, or is neutral to the premise (e.g., \texttt{"Premise: A woman is reading a book in the library. Hypothesis: A woman is swimming."} would be classified as a contradiction.) We sample 2000 questions with each of the three conditions as the ground truth, totaling to 12000 RTE questions across both datasets.

\textbf{4-Choice:\quad} We sample from the \textsl{\textbf{MMLU}} \citep{mmlupaper} collection of datasets. MMLU consists of 57 datasets spanning various topics, which can broadly grouped into four subject domains: \textsl{\textbf{(1) STEM}} (n=2400), \textsl{\textbf{(2) Social Science}} (n=2400), \textsl{\textbf{(3) Humanities}}  (n=4400), and \textsl{\textbf{(4) Others}} (n=2800), totaling 12000 questions.

\subsection{Models}

We test 12 LMs across 3 model families: Falcon3, Gemma3, and Llama3.1. Each family contains four LMs --- two pairs of smaller and larger models (e.g. 3B/10B for Falcon3 or 8B/70B for Llama3.1). Each such pair represents a base and an instruction-tuned version of an LM. For instance, the two pairs that comprise the Gemma3 family are \texttt{\{(Gemma3-12B, Gemma3-12B-IT), (Gemma3-27B, Gemma3-27B-IT)\}}. This model test set construction allows us to observe how bias may change with model size and instruction tuning while keeping the model architecture constant.

\subsection{Prompt Complexity}

We record model responses across three prompt formats varying in level of complexity: 

\begin{enumerate}
    \item \textbf{`Zeroshot'}: Only test question is presented,
    \item \textbf{`Instruction-only'}: One-line task instruction precedes the test question,
    \item \textbf{`Fewshot'}: One-line task instruction and two example question-answers precede the test question.
\end{enumerate}
Since the 3-choice datasets (SNLI and MNLI) inherently require instructions to define the objective and response format, we only test those datasets using the Instruction-only and Fewshot prompts. See Appendix \ref{sec:appendixprompts} for the full prompts.

\section{Experiment 1: Models Are Biased}

Figure \ref{fig:baselinemodelbiasbarchart} shows the label proportions for all models tested on all question-types, grouped by dataset, using the fewshot prompt format. The figure reveals that baseline model results (i.e. without applying correction) show varying degrees of inherent label bias. We can characterize some response behaviors across question-types and models. For yes-no datasets, we see a moderate bias towards answering `No' across all models. In NLI datasets, we see a general strong bias towards option `1', with the Gemma3-27B model showing the strongest preference for option `1' and the Llama3.1-8B model completely abstaining from responding with option `2'. In 4-choice datasets, we see relatively more uniform option preference across all models.

Across all models, we see that the bigger-size versions of models usually yield more uniform label distributions relative to their smaller counterpart. We see the same trend with the instruction-tuned version compared to the base version of the model. This supports the claim that using both bigger and instruction-tuned versions of models can inherently help reduce response bias.

\section{Experiment 2: Correction Improves Performance}

Figure \ref{fig:samecondscatterplots} shows the accuracy and bias changes for all models after applying our correction (we plot the median corrected accuracy and bias across 100 correction iterations using different calibration sets, each of size 100, using the fewshot prompt format). We pick one representative dataset from each question type for three plots, and the fourth plot shows the average changes across all ten datasets. 

We consistently see a reduction in bias value with either a higher or maintained accuracy across all models on all datasets. Models show the biggest accuracy increase in the 3-choice SNLI dataset compared to EWOK and MMLU-STEM. The final average scatterplot shows that the smaller-sized model pairs from the Gemma3 and Llama3.1 families had the biggest overall accuracy gains from applying the correction. This indicates that our bias correction can recover actual task performance that was lost due to learned response bias, and that it is especially effective for smaller-sized models.

\textbf{Significance of bias correction beyond accuracy improvement:} While the individual-dataset plots show low accuracy gains, we believe that the correction method is still worth applying for model evaluation purposes, because it allows for a fairer comparison of reasoning capabilities after removing the model's inherent label response bias. Removing the label bias and observing change in accuracy gives us important diagnostic information about label bias being a potential factor affecting the model's output performance. If model performance is in fact influenced by label bias, then removing the bias may uncover latent performance that was hidden as a result of the label priors. Conversely, if label bias is not a substantial driving factor for model performance, then removing label bias will not do much for improving accuracy, but it does still remove label bias as a potential confound to performance, allowing an evaluator to test for other biases or non-bias related problems affecting model performance. 

\section{Experiment 3: Comparison With Existing Methods}

Here we describe two existing LogProbs-based bias mitigation methods and perform them in our experimental setup for comparison with \texttt{RBCorr}.

\textbf{(1) Contextual Calibration (CC)} \citep{contextcalibpaper}:\quad This method is based on the idea that the model’s bias towards certain answers can be estimated by feeding in a content-free input, such as "N/A". For example, in a 2-shot prompt for a yes-no question where one example is a `Yes' and the other is a `No', we can append a third example with `N/A' as the question content, and extract the yes-no probabilities. The content-free nature of the third example is meant to reveal the model's contextual bias, which may change as the content or order of the preceding examples changes. The correction involves collecting and averaging model output probabilities using a few "content-free" input prompts (using "N/A", "[MASK]", and the empty string), and then using them to adjust test item probabilities via an affine matrix operation; this is equivalent to dividing the test item output probability with the mean content-free output probability.

\textbf{(2) Batch Calibration (BC)} \citep{batchcalibpaper}:\quad This method also focuses on mitigating the model's contextual bias, but instead of using content-free tokens, proposes to use a batch of items sampled from the dataset itself. The method first calculates a contextual prior over a batch, which is effectively a class-wise mean of the LogProbs, and subtracts it from the individual item LogProbs in that batch. This method of obtaining the correction term is meant to overcome CC's (and other methods') failure cases in multi-sentence classification. The batch mean calculated from the first processed batch is continuously refined using the items of incoming batches and applied to those items. 

Table \ref{tab:methodcomparetable} shows the bias and accuracy changes using the Llama3.1-8B and Llama3.1-8B-Instruct models on all datasets. Based on the results, no single method presents wins out across all datasets. Our method generally achieves the biggest accuracy gains for the 2-choice and 3-choice datasets, while Batch Calibration achieves the best gains for 4-choice datasets. Both BC and \texttt{RBCorr} show accuracy and bias changes in similar ranges and generally yield higher accuracy improvements compared to CC. The biggest improvement results in nearly 29\% of recovered accuracy (for the BABI dataset using \texttt{RBCorr}). In nearly all observed cases, \texttt{RBCorr} yields the biggest bias (TVD value) reduction of all three correction methods.

\section{Experiment 4: Correction Is Not Transferrable}

A unique aspect of \texttt{RBCorr} is that we calculate a single static set of mean LogProbs values for answer options in a dataset, which serve as `correction terms.' Once LogProbs values for any configuration (model, dataset, prompt format) are generated, the resulting bias correction terms can be applied to other LogProbs sets. This enables testing the \textbf{transferability} of correction terms across \textbf{models}, \textbf{datasets}, and \textbf{prompts}, effectively examining bias consistency across these modalities.

We test transfer efficacy by transferring correction terms across each modality independently (e.g., a cross-model correction will only source the term from other same-prompt, same-dataset runs). We conduct 100 correction runs with random calibration set sampling at a fixed size of 500 questions. Evidence of reliable correction in these transfer experiments should tell us whether response bias is most dependent on the model, dataset, or prompt being used for evaluation.

Table \ref{tab:transfertable} shows correction terms fail to transfer successfully across varying configurations, despite changing only one modality at a time and enforcing systematic constraints (only within-family transfer for cross-model, and only within-question-type transfer for cross-dataset). We define a "successful" transfer correction as one that achieves $\geq$80\% of both the TVD value reduction and accuracy gain that was yielded on applying the same-condition (i.e. non-transfer) correction on the tested target configuration.

Based on this success criteria, all three transfer modalities perform poorly, with cross-model and cross-prompt achieving success rates of only 10.42\%. Notably, Cross-dataset transfer shows marginally higher success (21.93\%), indicating bias behavior generalizes better by model and prompt configuration than by dataset. However, low success rates across all modalities demonstrate that condition-specific calibration is essential for reliable bias mitigation.

As an auxiliary analysis, Appendix \ref{sec:appendixheatmap} presents transfer results for three individual configurations. While not supporting general inference, it reveals transfer asymmetry: transferring from configuration A $\rightarrow$ B yields different effects than B $\rightarrow$ A.

\section{Related Works}

Several bias evaluation papers provide behavior characterizations across models and datasets. \citet{pezesh2023-optionorderingbias} demonstrate that the order in which options are presented can drastically impact model performance in ICL settings. \citet{salecha2024-llmsocialdesirabilitybias} demonstrates that LMs shift their responses to be more socially desirable when they are provided with enough questions to self-infer that they are being socially evaluated. \citet{tjuatja2024-llmhumanlikebias} develop a human-survey like dataset to test various LMs on five types of biases and show that models in general fail to reflect known human-like response patterns, highlighting the risk of using them as human proxy reasoners.

Our work falls more specifically into a smaller family of recent works that aim to mitigate bias using theoretically-motivated LogProbs manipulation methods. Two of these methods -- \citet{contextcalibpaper} Contextual Calibration and \citet{batchcalibpaper} Batch Calibration -- are used for comparison with our method, but there are other proposed methods as well, such as Contextual Calibration (DC) \citep{domaincalibpaper} and Prototypical Calibration (PC) \citep{protocalibpaper}. The DC method estimates a correction term by randomly sampling a sequence of tokens at average sentence length from an in-domain text set and averaging the model scores (i.e. probabilities for options) on that sequence multiple times, and finally dividing that mean score from the test-item probabilities during inference. The PC method is more sophisticated; it calibrates clusters of model output scores that are initially estimated using a Gaussian mixture model (GMM) and then refined using Expectation Maximization (EM)-based tuning. 

We decide to exclude comparison to them because the \cite{batchcalibpaper} work shows that BC performs better in general compared to PC and DC, making BC sufficient for performance comparison. We perform one additional auxiliary comparison between our method and the PriDe method introduced by \cite{pridepaper} --- this method is not similar in principle to the other methods in this family, since it scales overhead compute cost with correction efficacy, but still aims to correct label bias using prior estimation. PriDe comparison results and additional details are provided in Appendix \ref{sec:appendixpride}.
  
\section{Conclusion \& Future Work}

In this paper, we quantify LLM response bias in three kinds of closed-response questions using a experimental setup consisting of various datasets, models, and prompt schemes. We then propose a LogProbs-based bias correction method, \texttt{RBCorr}, that effectively reduces measured bias and improves accuracy, and compare it to similar existing methods to demonstrate its efficacy. Finally, we conduct an analysis to show that LogProbs values are highly specific to model, dataset, and prompt configurations, and bias estimations cannot reliably generalize over any of the three settings. 

\texttt{RBCorr} provides value both as a performance improver and an evaluation diagnostic technique. For performance improvement, a system can store a set of pre-computed calibration value for a given prompt and question type and apply them dynamically in online inference settings. The performance improvement we observe is particularly relevant to small and medium-sized LMs, offering an opportunity to use these lightweight cheap models for large-scale but relatively straightforward tasks (e.g., closed-form text labeling). For evaluation, our method helps uncover latent model performance and determine whether response bias hampers its performance on a task it could otherwise solve. In the age of skepticism of LM evaluations \citep{banerjee2024vulnerability, cao2025toward}, we posit that benchmark-based evaluations might still have value, as long as they are designed to measure general capabilities rather than specific task performance, maintain genuine train/test set separability, and are coupled with debiasing techniques that help uncover latent model performance.
 
Future work could explore using corrected probabilities as a tuning set to inherently debias the model's outputs, i.e., training a model to adjust its label token probabilities to align with previously-extracted and corrected probabilities, leading to better performance without having to apply post-hoc correction to the LogProbs results. We also point toward mechanistic interpretability approaches for measuring bias across model layers \citep[e.g., ][]{gupta2025llms}, tracing the origin of the bias in model weights, and correcting the bias at its source rather than at the last layer. Finally, we consider response bias to be a valuable test case for bias exploration and correction, which can then be extended to open-form responses and reasoning traces.

\section*{Limitations}

Our method is only applicable to open-source methods, for which LogProbs values of tokens are accessible. For closed models, one may try to estimate the bias by sampling responses with high temperature, but that method may be less precise. Additionally, we test transformer-only models up to 70B parameters in size; we do not report on the response behavior and correction efficacy on larger models or models based on other architectures, although we expect similar trends to hold. 

\section*{Acknowledgments}

We thank members of the LIT lab who provided feedback on earlier stages of this work. We also thank Han Zhou \citep{batchcalibpaper} for clarifying high-level details for implementing the Batch Calibration correction in our experiments. This research was supported in part through the computing resources provided by the Partnership for an Advanced Computing Environment (PACE) at the Georgia Institute of Technology.

\bibliography{rbcorr_acl}

\appendix

\begin{figure*}[t]
    \includegraphics[width=\linewidth]{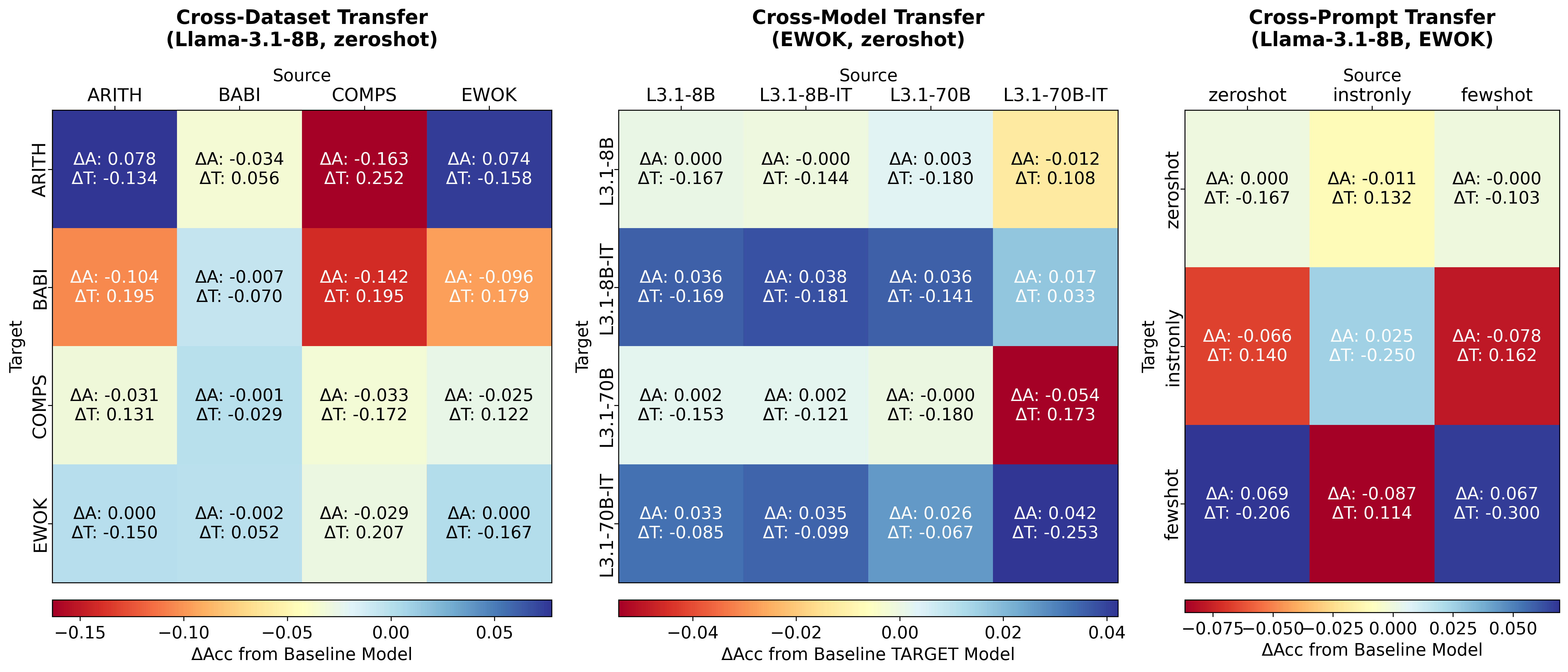}
    \caption{Heatmaps showing transfer correction performance for all three transfer modalities using three specific model-dataset-prompt setups. Gradient indicates accuracy change after applying correction relative to baseline model accuracy.}
    \label{fig:transferheatmaps}
\end{figure*}

\section{All Input Prompts}
\label{sec:appendixprompts}

Below are all the fewshot prompts used for all datasets. The instruction-only prompt format uses the same prompt scheme, simply without the examples, while the zeroshot prompt format has no text scaffolding and simply feeds in the test item from the dataset to the LM.

\subsection{ARITH}

\#INSTRUCTIONS\\Answer the following yes-no questions:\\\\\#EXAMPLE\\Question: Is 7 minus 9 equal to 4?\\Response: No\\\\\#EXAMPLE\\Question: Is 17 plus 15 equal to 32?\\Response: Yes\\\\\#EXAMPLE\\Question:

\subsection{BABI}

\#INSTRUCTIONS\\Answer the following yes-no questions:\\\\\#EXAMPLE\\Question: Marshall is in the car. Is Marshall in the building?\\Response: No\\\\\#EXAMPLE\\Question: Nathan is a pianist. Pianists like oranges. Does Nathan like oranges?\\Response: Yes\\\\\#EXAMPLE\\Question:

\subsection{COMPS}

\#INSTRUCTIONS\\Answer the following yes-no questions:\\\\\#EXAMPLE\\Question: Does a blueberry fire bullets?\\Response: No\\\\\#EXAMPLE\\Question: Does a turtle have a hard shell?\\Response: Yes\\\\\#EXAMPLE\\Question:

\subsection{EWOK}

\#INSTRUCTIONS\\Answer the following yes-no questions:\\\\\#EXAMPLE\\Question: Claire sees something that is fabric. Can Claire pour it?\\Response: No\\\\\#EXAMPLE\\Question: Sally pays salary to Harry. Is Sally Harry's boss?\\Response: Yes\\\\\#EXAMPLE\\Question:

\subsection{SNLI and MNLI}

\#INSTRUCTIONS\\Answer the following Recognizing Textual Entailment questions using a single digit. Entailment (0) implies the hypothesis is true given the premise. Neutral (1) implies the premise doesn't provide enough information to determine the hypothesis. Contradiction (2) implies the hypothesis is false given the premise.\\\\\#EXAMPLE\\Premise: A man is playing a guitar. Hypothesis: A person is making music.\\Response: 0\\\\\#EXAMPLE\\Premise: A woman is reading a book in the library. Hypothesis: A woman is swimming. \\Response: 2\\\\\#EXAMPLE:

\subsection{MMLU (All Domains)}

\#INSTRUCTIONS\\Answer the following multiple choice questions:\\\\\#EXAMPLE\\Question: What is the shape of the Earth?\\Options: (A) Cone, (B) Cube, (C) Sphere, (D) Cylinder\\Response: C\\\\\#EXAMPLE\\Question: What is the color of the sky?\\Options: (A) Red, (B) Blue, (C) Green, (D) Yellow\\Response: B\\\\\#EXAMPLE\\Question:

\section{RBCorr Transfer Correction -- Individual Configuration Heatmaps}
\label{sec:appendixheatmap}

We pick one specific configuration to illustrate the effects of \texttt{RBCorr} transfer correction in more granular detail. The configurations for each transfer modality are described below:

\begin{enumerate}
    \item \textbf{Cross-dataset}: Transfer among the 2-choice Yes-No datasets, using Llama3.1-8B as the model and zeroshot as the prompt level,
    \item \textbf{Cross-model}: Transfer among the Llama3.1 family of models, using EWOK as the dataset and zeroshot as the prompt level,
    \item \textbf{Cross-prompt}: Transfer among all three prompt levels, using EWOK as the dataset and Llama3.1-8B as the model.
\end{enumerate}

Figure \ref{fig:transferheatmaps} show the results of applying transfer correction in these setups. In general, transfer correction performs poorly and transfer efficacy is asymmetric. These specific transfer instances also allow us to make some generalizable hypotheses; for example, the cross-model results show that both the instruction-tuned Llama3.1 models are consistently good targets for cross-model correction, indicating that bias patterns for those models are especially invariant to the model and consequently more dependent on the dataset and prompt configurations. Such transfer analysis over other individual configuration setups may reveal other potential hypotheses to inform us of unknown model behavior traits.

\section{Comparing RBCorr with PriDe}
\label{sec:appendixpride}

\begin{table}[h]
\small
\renewcommand{\arraystretch}{1.2}
\setlength{\tabcolsep}{1.6pt}
\begin{tabular}{|c|ll|ll|}
\hline
\multirow{2}{*}{Model}                          & \multicolumn{2}{c|}{$\Delta \text{RStd} \%$ (MMLU)} & \multicolumn{2}{c|}{$\Delta \text{Acc.} \%$ (MMLU)} \\ \cline{2-5} 
 &
  \multicolumn{1}{c|}{\begin{tabular}[c]{@{}c@{}}Ours \\ $1\times$\end{tabular}} &
  \multicolumn{1}{c|}{\begin{tabular}[c]{@{}c@{}}PriDe\\ $1.15\times$\end{tabular}} &
  \multicolumn{1}{c|}{\begin{tabular}[c]{@{}c@{}}Ours\\$1\times$\end{tabular}} &
  \multicolumn{1}{c|}{\begin{tabular}[c]{@{}c@{}}PriDe\\ $1.15\times$\end{tabular}} \\ \hline
\multicolumn{1}{|l|}{\texttt{llama-2-7B}}                             & \multicolumn{1}{l|}{$-\textbf{77.55}$}   & $-76.09$   & \multicolumn{1}{l|}{$+\textbf{12.44}$}    & $+12.29$   \\ \hline
\multicolumn{1}{|l|}{\texttt{llama-2-chat-7B}}                        & \multicolumn{1}{l|}{$-\textbf{53.43}$}   & $-41.48$   & \multicolumn{1}{l|}{$+2.52$}    & $+9.58$    \\ \hline
\multicolumn{1}{|l|}{\texttt{llama-2-13B}}      & \multicolumn{1}{l|}{$-58.63$}   & $-69.33$   & \multicolumn{1}{l|}{$+\textbf{3.46}$}    & $+1.97$    \\ \hline
\multicolumn{1}{|l|}{\texttt{llama-2-chat-13B}} & \multicolumn{1}{l|}{$-34.78$}   & $-42.66$   & \multicolumn{1}{l|}{$+\textbf{2.97}$}     & $+2.68$    \\ \hline
\end{tabular}
\caption{Percentage changes in RStd (bias) and accuracy before and after applying LogProbs bias correction (ours) vs. PriDe. Higher percent decreases for RStd imply greater debiasing effect. Cases where our method results in larger improvement are highlighted.}
\label{tab:oursvspridetable}
\end{table} 

We compare our method's efficacy to the PriDe method from \cite{pridepaper}. Since their paper provides comprehensive debiasing results, including results from the Llama2 model family on the MMLU datasets, we run our correction method on the same models and datasets in order to directly use their provided results for comparison. Since their method scales efficacy with compute cost, while our method requires zero extra compute, we compare our method's results to the PriDe variation with the lowest cost, which carries a 15\% (i.e. $1.15\times$) overhead compute. We also quantify our method's efficacy using the same bias metric \cite{pridepaper} uses, i.e., RSTD or standard deviation over recalls. 

Table \ref{tab:oursvspridetable} shows the accuracy and bias value changes achieved by performing \texttt{RBCorr} and PriDe on the same models and datasets. We observe that our method achieves comparable effects for both bias reduction and accuracy improvement while having zero overhead compute. These results are not strongly generalizable since we compare only a single model family, but show promise for simple LogProbs-based manipulations as a competent way to mitigate model bias.

\end{document}